\documentclass[10pt,twocolumn,letterpaper]{article}

\usepackage{cvpr}
\usepackage{times}
\usepackage{epsfig}
\usepackage{graphicx}
\usepackage{amsmath}
\usepackage{amssymb}
\usepackage{paralist}
\usepackage{multirow}
\usepackage{booktabs}
\usepackage[normalem]{ulem}
\useunder{\uline}{\ul}{}

\usepackage[pagebackref=true,breaklinks=true,letterpaper=true,colorlinks,bookmarks=false]{hyperref}

\cvprfinalcopy
\def\cvprPaperID{****}

\ifcvprfinal\fi
\begin{document}

\newcommand{\param}{\mathbf{w}}
\newcommand{\keypt}{\mathbf{x}}
\newcommand{\keypts}{\mathcal{X}}
\newcommand{\matches}{\mathcal{M}}
\newcommand{\match}{\mathbf{m}}

\newcommand{\descr}[1]{\mathbf{d}(#1)}

\newcommand{\loss}{\ell}
\newcommand{\Loss}{\mathcal{L}}

\newcommand{\est}[1]{\hat{#1}}
\newcommand{\gt}[1]{{#1}^*}

\newcommand{\expectation}[2]{\mathbb{E}_{#1}\left[ #2 \right]}
\newcommand{\derv}[1]{\frac{\partial}{\partial #1}}

\vspace{-0.5cm}
\title{Reinforced Feature Points: \\Optimizing Feature Detection and Description for a High-Level Task\vspace{-0.4cm}}

\author{Aritra Bhowmik$^1$, Stefan Gumhold$^1$, Carsten Rother$^2$, Eric Brachmann$^2$\\
$^1$ TU Dresden, $^2$ Heidelberg University
}

\maketitle

\begin{abstract}
We address a core problem of computer vision: Detection and description of 2D feature points for image matching.
For a long time, hand-crafted designs, like the seminal SIFT algorithm, were unsurpassed in accuracy and efficiency.
Recently, learned feature detectors emerged that implement detection and description using neural networks. 
Training these networks usually resorts to optimizing low-level matching scores, often pre-defining sets of image patches which should or should not match, or which should or should not contain key points.
Unfortunately, increased accuracy for these low-level matching scores does not necessarily translate to better performance in high-level vision tasks.
We propose a new training methodology which embeds the feature detector in a complete vision pipeline, and where the learnable parameters are trained in an end-to-end fashion. 
We overcome the discrete nature of key point selection and descriptor matching using principles from reinforcement learning.
As an example, we address the task of relative pose estimation between a pair of images.
We demonstrate that the accuracy of a state-of-the-art learning-based feature detector can be increased when trained for the task it is supposed to solve at test time.
Our training methodology poses little restrictions on the task to learn, and works for any architecture which predicts key point heat maps, and descriptors for key point locations.
\end{abstract}
\vspace{-0.5cm}

\begin{figure}
\begin{center}
\includegraphics[width=\linewidth]{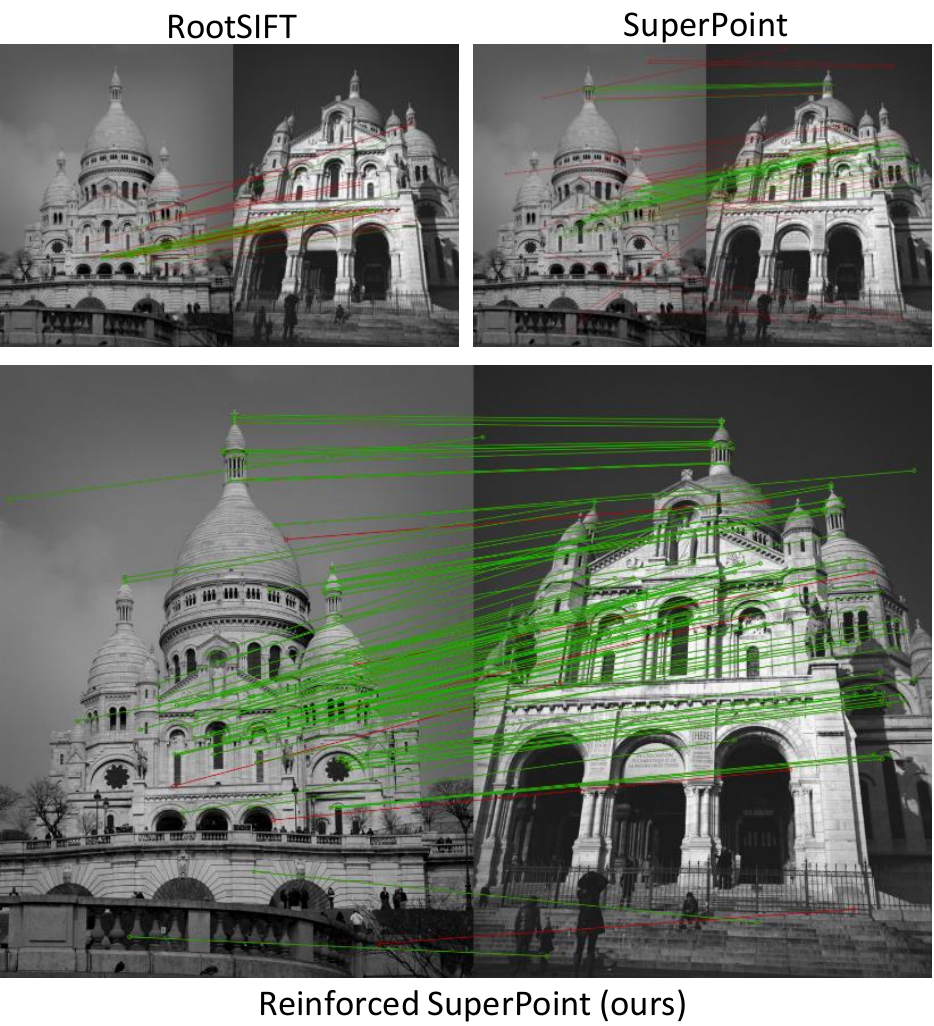}
\end{center}
    \vspace{-0.3cm}
    \caption{We show the results of estimating the relative pose (essential matrix) between two images using RootSIFT \cite{rootsift12} (top left) and SuperPoint \cite{detone18superpoint} (top right). Our Reinforced SuperPoint (bottom), utilizing \cite{detone18superpoint} within our proposed training schema, achieves a clearly superior result. Here the inlier matches wrt.~the ground truth essential matrix are drawn in green, outliers in red.}
    \vspace{-0.5cm}
\label{fig:teaser}
\end{figure}

\section{Introduction}
\vspace{-0.1cm}

Finding and matching sparse 2D feature points across images has been a long-standing problem in computer vision \cite{Harris88alvey}.
Feature detection algorithms enable the creation of vivid 3D models from image collections \cite{reconworld15, visualsfm13,schoenberger2016sfm}, building maps for robotic agents \cite{orbslam2015,orbslam22017}, recognizing places \cite{schindler2007city,li2010location,Noh2017Delf} and precise locations  \cite{li2016worldwide,Toft2018ECCV,sattler2016efficient} as well as recognizing objects \cite{Lowesift,nister06scalable,Philbin07,rootsift12,vlad2013}.
Naturally, the design of feature detection and description algorithms, subsumed as feature detection in the following, has received tremendous attention in computer vision research since its early days.
Although invented three decades ago, the seminal SIFT algorithm \cite{Lowesift} remains the gold standard feature detection pipeline to this day.

With the recent advent of powerful machine learning tools, some authors replace classical, feature-based vision pipelines by neural networks \cite{kendall2015convolutional, demon17, relocnet2018}. 
However, independent studies suggest that these learned pipelines have not yet reached the accuracy of their classical counterparts \cite{schonberger2017comparative,Sattler2018changing, goodcorr18, Sattler2019limitations}, due to limited generalization abilities.
Alternatively, one prominent strain of current research aims to keep the concept of sparse feature detection but replaces hand-crafted designs like SIFT \cite{Lowesift} with data-driven, learned representations. 
Initial works largely focused on learning to compare image patches to yield expressive feature descriptors \cite{han2015matchnet, tian2017l2net, Balntas2016triplet, Mishchuk2017hardnet, luo2018geodesc, sosnet2019cvpr}. Fewer works attempt to learn feature detection \cite{Cieslewski19threedv, BarrosoLaguna2019keynet} or a complete architecture for feature detection and description \cite{lift2016, Noh2017Delf, detone18superpoint}.

Training of these methods is usually driven by optimizing low-level matching scores inspired by metric learning \cite{Balntas2016triplet} with the necessity to define ground truth correspondences between patches or images.
When evaluated on low-level matching benchmarks like H-Patches \cite{hpatches2017hpatches}, such methods regularly achieve highly superior scores compared to a SIFT baseline.
H-Patches \cite{hpatches2017hpatches} defines sets of matching image patches that undergo severe illumination and viewpoint changes.
However, the increased accuracy in such matching tasks does not necessarily translate to increased accuracy in high-level vision pipelines. 
For example, we show that the state-of-the-art learned SuperPoint detector \cite{detone18superpoint}, while highly superior to SIFT \cite{Lowesift} on H-Patches \cite{hpatches2017hpatches}, does not reach SIFT's capabilities when estimating an essential matrix for an image pair.
Similar observations were reported in earlier studies, where the supposedly superior learned LIFT detector \cite{lift2016} failed to produce richer reconstructions than SIFT \cite{Lowesift} in a structure-from-motion pipeline \cite{schonberger2017comparative}.

Some authors took notice of the discrepancy between low-level training and high-level performance, and developed training protocols that mimic properties of high-level vision pipelines. 
Lua \etal \cite{luo2018geodesc} perform hard negative mining of training patches in a way that simulates the problem of self-similarity when matching at the image level. 
Revaud \etal \cite{Revaud2019r2d2} train a detector to find few but reliable key points.
Similarly Cieslewski \etal \cite{Cieslewski19threedv} learn to find key points with high probability of being inliers in robust model fitting.

In this work, we take a more radical approach. Instead of hand-crafting a training procedure that emulates aspects of high-level vision pipelines, we embed the feature detector in a complete vision pipeline during training.
Particularly, our pipeline addresses the task of relative pose estimation, a central component in camera re-localization, structure-from-motion or SLAM. 
The pipeline incorporates key point selection, descriptor matching and robust model fitting. 
We do not need to pre-define ground truth correspondences, dispensing with the need for hard-negative mining. 
Furthermore, we do not need to speculate whether it is more beneficial to find many matches or few, reliable matches.
All these aspects are solely guided by the task loss, \ie by minimizing the relative pose error between two images.

Key point selection and descriptor matching are discrete operations which cannot be directly differentiated. 
However, since many feature detectors predict key point locations as heat maps, we can reformulate key point selection as a sampling operation.
Similarly, we lift feature matching to a distribution where the probability of a match stems from its descriptor distance. 
This allows us to apply principles from reinforcement learning \cite{rlbook} to directly optimize a high-level task loss. 
Particularly, all operations after the feature matching stage, \eg robust model fitting, do not need to be differentiable since they only provide a reward signal for learning.
In summary, our training methodology puts little restrictions on the feature detection architecture or the vision task to be optimized for.

We demonstrate our approach using the SuperPoint detector \cite{detone18superpoint}, which regularly ranks among top methods in independent evaluations \cite{Dusmanu2019d2net, BarrosoLaguna2019keynet, Revaud2019r2d2}.
We train SuperPoint for the task of relative pose estimation by robust fitting of the essential matrix.
For this task, our training procedure closes the gap between SuperPoint and a state-of-the-art SIFT-based pipeline, see Fig.~\ref{fig:teaser} for a comparison of results.

\noindent We summarize our main contributions:

\begin{compactitem}
\item A new training methodology which allows for learning a feature detector and descriptor, embedded in a complete vision pipeline, to optimize its performance for a high-level vision task.
\item We apply our method to a state-of-the-art architecture, Superpoint \cite{detone18superpoint}, and train it for the task of relative pose estimation. 
\item After training, SuperPoint \cite{detone18superpoint} reaches, and slightly exceeds, the accuracy of SIFT \cite{Lowesift} which previously achieved best results for this task.
\end{compactitem}
\vspace{-0.2cm}

\section{Related Work}
\vspace{-0.1cm}

Of all hand-crafted feature detectors, SIFT \cite{Lowesift} stands out for its long lasting success.
SIFT finds key point locations as a difference-of-Gaussian filter response in the scale space of an image, and describes features using histograms of oriented gradients \cite{dalal2005hog}.
Arandjelovic and Zisserman \cite{rootsift12} improve the matching accuracy of SIFT by normalizing its descriptor, also called RootSIFT.
Other hand-crafted feature detectors improve efficiency for real-time applications while sacrificing as little accuracy as possible \cite{bay06surf,rublee2011orb}.

MatchNet \cite{han2015matchnet} is an early example of learning to compare image patches using a patch similarity network.
The reliance on a network as a similarity measure prevents the use of efficient nearest neighbor search schemes.
L2-Net \cite{tian2017l2net}, and subsequent works, instead learn patch descriptors to be compared using the Euclidean distance.
Balntas \etal \cite{Balntas2016triplet} demonstrated the advantage of using a triplet loss for descriptor learning over losses defined on pairs of patches only.
A triplet combines two matching and one non-matching patch, and the triplet loss optimizes relative distances within a triplet.
HardNet \cite{Mishchuk2017hardnet} employs a ``hardest-in-batch" strategy when assembling triplets for training, \ie for each matching patch pair, they search for the most similar non-matching patch within a mini-batch.
GeoDesc \cite{luo2018geodesc} constructs mini-batches for training that contain visually similar but non-matching patch pairs to mimic the problem of self-similarity when matching two images.
SOSNet \cite{sosnet2019cvpr} uses second order similarity regularization to enforce a structure of the descriptor space that leads to well separated clusters of similar patches.

Learning feature \emph{detection} has also started to attract attention recently.
ELF \cite{Benbihi2019elf} shows that feature detection can be implemented using gradient tracing within a pre-trained neural network.
Key.Net \cite{BarrosoLaguna2019keynet} combines hand-crafted and learned filters to avoid overfitting.
The detector is trained using a repeatability objective, \ie finding the same points in two related images, synthetically created by homography warping.
SIPs \cite{Cieslewski19threedv} learns to predict a pixel-wise probability map of inlier locations as key points, inlier being a correspondence which can be continuously tracked throughout an image sequence by an off-the-shelf feature tracker.

LIFT \cite{lift2016} was the first, complete learning-based architecture for feature detection \emph{and} description.
It rebuilds the main processing steps of SIFT with neural networks, and is trained using sets of matching and non-matching image patches extracted from structure-from-motion datasets.
DELF \cite{Noh2017Delf} learns detection and description for image retrieval, where coarse key point locations emerge by training an attention layer on top of a dense descriptor tensor.
D2-Net \cite{Dusmanu2019d2net} implements feature detection and description by searching for local maxima in the filter response map of a pre-trained CNN. 
R2D2 \cite{Revaud2019r2d2} proposes a learning scheme for identifying feature locations that can be matched uniquely among images, avoiding repetitive patterns.


All mentioned learning-based works design training schemes that emulate difficult conditions for a feature detector when employed for a vision task.
Our work is the first to directly embed feature detection and description in a complete vision pipeline for training where all real-world challenges occur, naturally.
On a similar note, KeypointNet \cite{suwajanakorn2018discovery} describes a differentiable pipeline that automatically discovers category-level key points for the task of relative pose estimation.
However, \cite{suwajanakorn2018discovery} does not consider feature description nor matching.
In recent years, Brachmann \etal described a differentiable version of RANSAC (DSAC) \cite{brachmann2017dsac, brachmann2018dsac++} to learn a camera localization pipeline end-to-end.
Similar to DSAC, we derive our training objective from policy gradient \cite{rlbook}.
However, by formulating feature detection and matching via sampling we do not require gradients of RANSAC, and hence we do not utilize DSAC.

We realize our approach using the SuperPoint \cite{detone18superpoint} architecture, a fully convolutional CNN for feature detection and description, pre-trained on synthetic and homography-warped real images.
In principle, our training scheme can be applied to architectures other than SuperPoint, like LIFT \cite{lift2016} or R2D2 \cite{Revaud2019r2d2}, and also to separate networks for feature detection and description.
\vspace{-0.2cm}

\begin{figure*}[ht!]
\begin{center}
\vspace{-0.5cm}
\includegraphics[width=\linewidth]{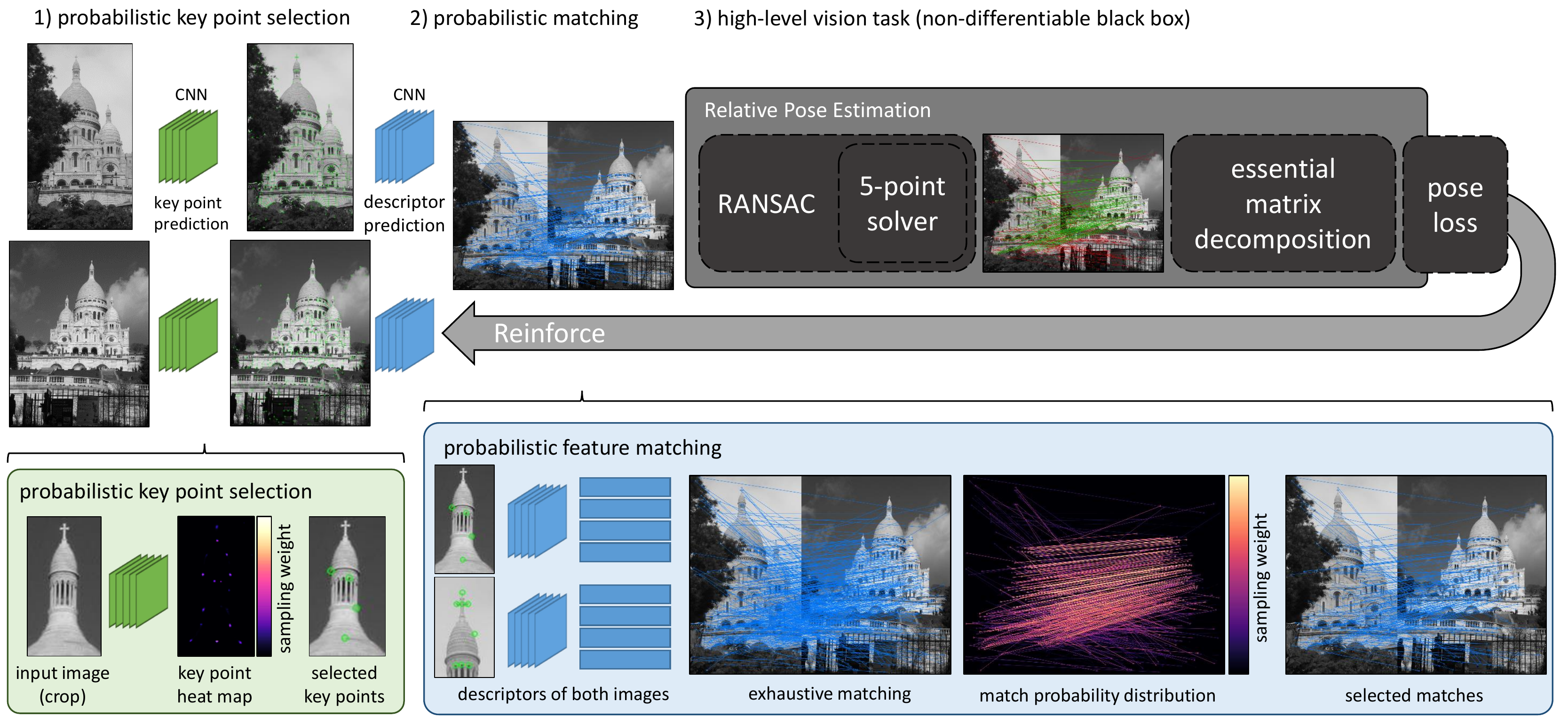}
\end{center}
   \vspace{-0.3cm}
   \caption{\textbf{Method. Top:} Our training pipeline consists of probabilistic key point selection and probabilistic feature matching. Based on established feature matches, we solve for the relative pose and compare it to the ground truth pose. We treat the vision task as a (potentially non-differentiable) black box. It provides an error signal, used to reinforce the key point and matching probabilities. Based on the error signal both CNNs (green and blue) are updated, making a low loss more likely. \textbf{Bottom Left:} We sample key points according to the heat map predicted by the detector. \textbf{Bottom Right:} We implement probabilistic matching by, firstly, doing an exhaustive matching between all key points, secondly, calculating a probability distribution over all matches depending on their descriptor distance, and, thirdly, sampling a subset of matches. We pass only this subset of matches to the black box estimator. }
   \label{fig:system}
   \vspace{-0.3cm}
\end{figure*}

\section{Method}
\vspace{-0.1cm}

As an example of a  high-level vision task, we estimate the relative transformation $T = (R,\mathbf{t})$, with rotation $R$ and translation $\mathbf{t}$, between two images $I$ and $I'$.
We solve the task using sparse feature matching. 
We determine 2D key points $\keypt_i$ indexed by $i$, and compute a descriptor vector $\descr{\keypt_i}$ for each key point.
Using nearest neighbor matching in descriptor space, we establish a set of tentative correspondences  $\match_{ij} = (\keypt_i,\keypt'_j)$ between images $I$ and $I'$.
We solve for the relative pose based on these tentative correspondences by robust fitting of the essential matrix \cite{mutliview2004}.
We apply a robust estimator like RANSAC \cite{ransac1981} with a 5-point solver \cite{nister2004fivepoint} to find the essential matrix which maximises the inlier count among all correspondences.
An inlier is defined as a correspondence with a distance to the closest epipolar line below a threshold \cite{mutliview2004}.
Decomposition of the essential matrix yields an estimate of the relative transformation $\est{T}$.

We implement feature detection using two networks: a detection network and a description network.
In practice, we use a joint architecture, SuperPoint\cite{detone18superpoint}, where most weights are shared between detection and description.
The main goal of this work is to optimize the learnable parameters $\param$ of both networks such that their accuracy for the vision task is enhanced.
For our application, the networks should predict key points and descriptors such that the relative pose error between two images is minimized.
Key point selection and feature matching are discrete, non-differentiable operations. Therefore, we cannot directly propagate gradients of our estimated transformation $\est{T}$ back to update the network weights, as in standard supervised learning.
Components of our vision pipeline, like the robust estimator (\eg RANSAC \cite{ransac1981}) or the minimal solver (\eg the 5-point solver \cite{nister2004fivepoint}) might also be non-differentiable. 
To optimize the neural network parameters for our task, we apply principles from reinforcement learning \cite{rlbook}.
We formulate feature detection and matching as probabilistic actions where the \emph{probability} of taking an action, \ie selecting a key point, or matching two features, depends on the output of the neural networks. 
During training, we sample different instantiations of key points and their matchings based on the probability distributions predicted by the neural networks.
We observe how well these key points and their matching perform in the vision task, and adjust network parameters $\param$ such that an outcome with low loss becomes more probable.
We show an overview of our approach in Fig.~\ref{fig:system}.

In the following, we firstly describe how to reformulate key point selection and feature matching as probabilistic actions.
Thereafter, we formulate our learning objective, and how to efficiently approximate it using sampling.

\subsection{Probabilistic Key Point Selection}

We assume that the detection network predicts a key point heat map $f(I; \param)$ for an input image, as is common in many architectures \cite{lift2016, detone18superpoint, Revaud2019r2d2, Cieslewski19threedv}.
Feature locations are usually selected from $f(I; \param)$ by taking all local maxima combined with local non-max suppression.

To make key point selection probabilistic, we instead interpret the heat map as a probability distribution over key point locations $f(I; \param) = P(\keypt;\param)$ parameterized by the network parameters $\param$.
We define a set of $N$ key points for image $I$  as $\keypts = \{\keypt_i\}$ sampled independently according to 
\begin{equation}
\label{eq:keypts}
\vspace{-0.1cm}
P(\keypts;\mathbf{w}) = \prod_{i=1}^N P(\keypt_i;\param),
\vspace{-0.1cm}
\end{equation}
see also Fig.~\ref{fig:system}, bottom left. 
Similarly, we define $\keypts'$ for image $I'$.
We give the joint probability of sampling key points independently in each image as 
\begin{equation}
\label{eq:keyptsjoint}
\vspace{-0.1cm}
P(\keypts,\keypts';\mathbf{w}) = P(\keypts;\mathbf{w})P(\keypts';\mathbf{w}).
\end{equation}

\subsection{Probabilistic Feature Matching}

We assume that a second description network predicts a feature descriptor $\descr{\keypt; \param}$ for a given key point $\keypt$.
To simplify notation, we use $\param$ to denote the learnable parameters associated with feature detection \emph{and} description.
We define a feature match as a pairing of one key point from image $I$ and image $I'$, respectively: $\match_{ij} = (\keypt_i,\keypt'_j)$.
We give the probability of a match between two key points $\keypt_i$ and $\keypt'_j$ as a function of their descriptor distance,
\def\-{\raisebox{.75pt}{-}}
\begin{equation}
\label{eq:match}
P(\match_{ij}|\keypts,\keypts';\param) = \frac{\exp[\-||\descr{\keypt_i;\param}\-\descr{\keypt'_j;\param}||]}{\sum_{\match_{kk'}}\exp[\-||\descr{\keypt_k;\param}\-\descr{\keypt'_{k'};\param}||]}.
\end{equation}
Note that the matching probability is conditioned on the sets of key points which we selected in an earlier step.
The matching distribution is normalized using all possible matches $\match_{kk'} = (\keypt_k,\keypt'_{k'})$ with $\keypt_k \in \keypts$ and $\keypt'_{k'} \in \keypts'$.
The matching distribution assigns a low probability to a match, if the associated key points have very different descriptors. 
To increase the probability of a (good) match during training, the network has to reduce the associated descriptor distance relative to all other matches for the image pair.

We define a complete set of $M$ matches $\matches = \{\match_{ij}\}$ between $I$ and $I'$ sampled independently according to 
\begin{equation}
\label{eq:matches}
P(\matches|\keypts,\keypts';\param) = \prod_{\match_{ij}\in\matches} P(\match_{ij}|\keypts,\keypts';\param).
\end{equation}

\vspace{-0.3cm}
\subsection{Learning Objective}

We learn network parameters $\param$ in a supervised fashion, \ie we assume to have training data of the form $(I,I',\gt{T})$ with ground truth transformation $\gt{T}$.
Note that we do \emph{not} need ground truth key point locations $\keypts$ or ground truth image correspondences $\matches$.

Our learning formulation is agnostic to the implementation details of how the vision task is solved using tentative image correspondences $\matches$.
We treat the exact processing pipeline after the feature matching stage as a black box which produces only one output: a loss value $\loss(\matches,\keypts,\keypts')$, which depends on the key points $\keypts$ and $\keypts'$ that we selected for both images, and the matches $\matches$ that we selected among the key points.
For relative pose estimation, calculating $\loss$ entails robust fitting of the essential matrix, its decomposition to yield an estimated relative camera transformation $\est{T}$, and its comparison to a ground truth transformation $\gt{T}$.
We only require the loss value itself, not its gradients.

Our training objective aims at reducing the expected task loss when sampling key points and matches according to the probability distributions parameterized by the learnable parameters $\param$:
\begin{equation}
\label{eq:trainingobj}
\vspace{-0.1cm}
\begin{split}
\Loss(\param) & = \expectation{\matches,\keypts,\keypts'\sim P(\matches,\keypts,\keypts';\mathbf{w})}{\loss(\matches,\keypts,\keypts')} \\
& = \mathbb{E}_{\keypts,\keypts'\sim P(\keypts,\keypts';\mathbf{w})} \expectation{\matches\sim P(\matches|\keypts,\keypts';\mathbf{w})}{\loss(\cdot)},
\end{split}
\vspace{-0.1cm}
\end{equation}
where we abbreviate $\loss(\matches,\keypts,\keypts')$ to $\loss(\cdot)$.
We split the expectation in key point selection and match selection.
Firstly, we select key points $\keypts$ and $\keypts'$ according to the heat map predictions of the detection network $P(\keypts,\keypts';\mathbf{w})$ (see Eq.~\ref{eq:keypts} and \ref{eq:keyptsjoint}).
Secondly, we select matches among these key points according to a probability distribution $P(\matches|\keypts,\keypts';\mathbf{w})$ calculated from descriptor distances (see Eq.~\ref{eq:match} and \ref{eq:matches}).

Calculating the expectation and its gradients exactly would necessitate summing over all possible key point sets, and all possible matchings, which is clearly infeasible.
To make the calculation tractable, we assume that the network is already initialized, and makes sensible predictions that we aim at optimizing further for our task.
In practice, we take an off-the-shelf architecture, like SuperPoint \cite{detone18superpoint}, which was trained on a low-level matching task.
For such an initialized network, we observe the following properties:
\begin{compactenum}
\item Heat maps predicted by the feature detector are sparse. The probability of selecting a key point is zero at almost all image pixels (see Fig.~\ref{fig:system} bottom, left). Therefore, only few image locations have an impact on the expectation.
\item Matches among unrelated key points have a large descriptor distance. Such matches have a probability close to zero, and no impact on the expectation.
\end{compactenum}
Observation 1) means, we can just sample from the key point heat map, and ignore other image locations.
Observation 2) means that for the key points we selected, we do not have to realise a complete matching of all key points in $\keypts$ to all key points in $\keypts'$.
Instead, we rely on a k-nearest-neighbour matching with some small $k$.
All nearest neighbours beyond $k$ likely have large descriptor distances, and hence near zero probability.
In practice, we found no advantage in using a $k>1$ which means we can do a normal nearest neighbour matching during training when calculating $P(\matches|\keypts,\keypts';\mathbf{w})$ (see Fig.~\ref{fig:system} bottom, right).

We update the learnable parameters $\param$ according to the gradients of Eq.~\ref{eq:trainingobj}, following the classic REINFORCE algorithm \cite{Williams1992reinforce} of Williams:
\begin{equation}
\label{eq:traingrad}
\vspace{-0.1cm}
\begin{split}
 \derv{\param} &\Loss(\param) = \\
 & \expectation{\keypts,\keypts'}{\expectation{\matches|\keypts,\keypts'}{\loss(\cdot)}\derv{\param}{\log P(\keypts,\keypts';\param)}} \\
  + &\expectation{\keypts,\keypts'}{\expectation{\matches|\keypts,\keypts'}{\loss(\cdot)\derv{\param}{\log P(\matches|\keypts,\keypts';\param)}}}.
\end{split}
\vspace{-0.1cm}
\end{equation}
Note that we only need to calculate the gradients of the log probabilities of key point selection and feature matching.
We approximate the expectations in the gradient calculation by sampling.
We approximate $\mathbb{E}_{\keypts,\keypts'}$ by drawing $n_\keypts$ samples $\est{\keypts},\est{\keypts}'\sim P(\keypts,\keypts';\param)$.
For a given key point sample, we approximate $\mathbb{E}_{\matches|\keypts,\keypts'}$ by drawing $n_\matches$ samples $\est{\matches}\sim P(\matches|\est{\keypts},\est{\keypts}';\param)$.
For each sample combination, we run the vision pipeline and observe the associated task loss $\loss$.
To reduce the variance of the gradient approximation, we subtract the mean loss over all samples as a baseline \cite{rlbook}.
We found a small number of samples for $n_\keypts$ and $n_\matches$ sufficient for the pipeline to converge.
\vspace{-0.2cm}

\section{Experiments}
\vspace{-0.1cm}

We train the SuperPoint \cite{detone18superpoint} architecture for the task of relative pose estimation, and report our main results in Sec.~\ref{sec:exp:relpose}.
Furthermore, we analyse the impact of reinforcing SuperPoint for relative pose estimation on a low-level matching benchmark (Sec.~\ref{sec:exp:hpatches}), and in a structure-from-motion task (Sec.~\ref{sec:exp:sfm}).
\vspace{-0.1cm}

\subsection{Relative Pose Estimation}
\label{sec:exp:relpose}

\paragraph{Network Architecture.} 
SuperPoint \cite{detone18superpoint} is a fully-convolutional neural network which processes full-sized images.
The network has two output heads: one produces a heat map from which key points can be picked, and the other head produces 256-dimensional descriptors as a dense descriptor field over the image.
The descriptor output of SuperPoint fits well into our training methodology, as we can look up descriptors for arbitrary image locations without doing repeated forward passes of the network.
Both output heads share a common encoder which processes the image and reduces its dimensionality, while the output heads act as decoders.
We use the network weights provided by the authors as an initialization.
\vspace{-0.3cm}

\paragraph{Task Description.} 
We calculate the relative camera pose between a pair of images by robust fitting of the essential matrix. 
We show an overview of the processing pipeline in Fig.~\ref{fig:system}.
The feature detector produces a set of tentative image correspondences. 
We estimate the essential matrix using the 5-point algorithm \cite{nister2004fivepoint} in conjunction with a robust estimator.
For the robust estimator, we conducted experiments with a standard RANSAC \cite{ransac1981} estimator, as well as with the recent NG-RANSAC \cite{Brachmann2019ngransac}.
NG-RANSAC uses a neural network to suppress outlier correspondences, and to guide RANSAC sampling towards promising candidates for the essential matrix.
As a learning-based robust estimator, NG-RANSAC is particularly interesting in our setup, since we can refine it in conjunction with SuperPoint during end-to-end training.
\vspace{-0.3cm}

\paragraph{Datasets.}
To facilitate comparison to other methods, we follow the evaluation protocol of Yi \etal \cite{goodcorr18} for relative pose estimation.
They evaluate using a collection of 7 outdoor and 16 indoor datasets from various sources \cite{Strecha2008calib, reconworld15, xiao2013sun3d}.
One outdoor scene and one indoor scene serve as training data, the remaining 21 scenes serve as test set.
All datasets come with co-visibility information for the selection of suitable image pairs, and ground truth poses.
\vspace{-0.4cm}

\paragraph{Training Procedure.}
We interpret the output of the detection head of SuperPoint as a probability distribution over key point locations.
We sample 600 key points for each image, and we read out the descriptor for each key point from the descriptor head output.
Next, we perform a nearest neighbour matching between key points, accepting only matches of mutual nearest neighbors in both images.
We calculate a probability distribution over all the matches depending on their descriptor distance (according to Eq.~\ref{eq:match}).
We randomly choose $50\%$ of all matches from this distribution for the relative pose estimation pipeline.
We fit the essential matrix, and estimate the relative pose up to scale.
We measure the angle between the estimated and ground truth rotation, as well as, the angle between the estimated and the ground truth translation vector.
We take the maximum of both angles as our task loss $\loss$. 
For difficult image pairs, essential matrix estimation can fail, and the task loss can be very large.
To limit the influence of such large losses, we apply a square root soft clamping \cite{Brachmann2019ngransac} of the loss after a value of $25^\circ$, and a hard clamping after a value of $75^\circ$.

To approximate the expected task loss $\Loss(\param)$ and its gradients in Eq.~\ref{eq:trainingobj} and Eq.~\ref{eq:traingrad}, we draw key points $n_\keypts=3$ times, and, for each set of key points, we draw $n_\matches=3$ sets of matches.
Therefore, for each training iteration, we run the vision pipeline $9$ times, which takes 1.5s to 2.1s on a single Tesla K80 GPU, depending on the termination of the robust estimator.
We train using the Adam \cite{adam2014} optimizer and a learning rate of $10^{-7}$ for 150k iterations which takes approximately 60 hours.
Our training code is based on PyTorch \cite{paszke2017automatic} for SuperPoint \cite{detone18superpoint} integration and learning, and on OpenCV \cite{opencv_library} for estimating the relative pose.
We will make our source code publicly available to ensure reproducibility of our approach.
\vspace{-0.4cm}

\paragraph{Test Procedure.}
For testing, we revert to a deterministic procedure for feature detection, instead of doing sampling.
We select the strongest 2000 key points from the detector heat map using local non-max suppression. 
We remove very weak key point with a heat map value below $0.00015$.
We do a nearest neighbor matching of the corresponding feature descriptors, and keep all matches of mutual nearest neighbors.
We adhere to this procedure for SuperPoint before and after our training, to ensure comparability of the results.
\vspace{-0.4cm}

\begin{figure*}
\begin{center}
\vspace{-0.4cm}
\includegraphics[width=1\linewidth]{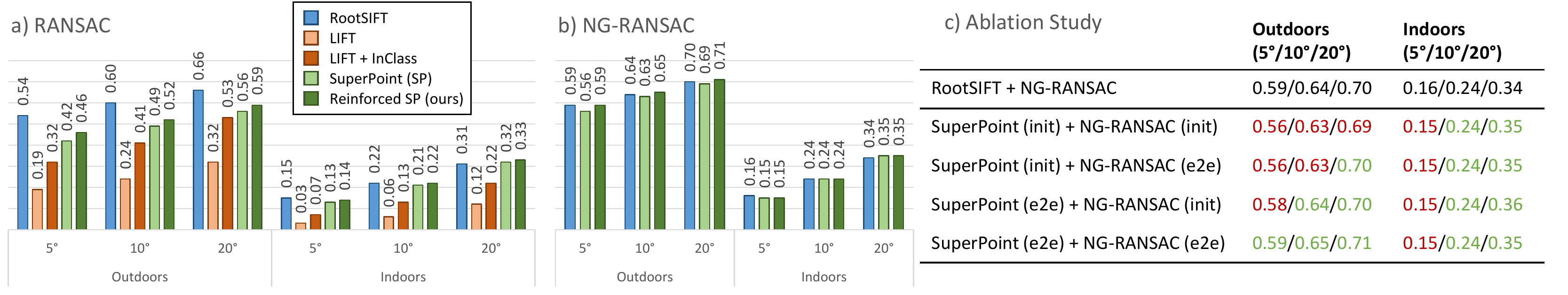}
\end{center}
\vspace{-0.4cm}
   \caption{\textbf{Relative Pose Estimation. a)} AUC of the relative pose error using a RANSAC estimator for the essential matrix. Results of RootSIFT as reported in \cite{Brachmann2019ngransac}, results of LIFT as reported in \cite{goodcorr18}. \textbf{b)} AUC using NG-RANSAC \cite{Brachmann2019ngransac} as robust estimator. \textbf{c)} For our best result, we show the impact of training SuperPoint vs.~NG-RANSAC end-to-end. \emph{Init.}~for SuperPoint means weights provided by Detone \etal \cite{detone18superpoint}, \emph{init.}~for NG-RANSAC means training according to Brachmann and Rother \cite{Brachmann2019ngransac} for SuperPoint. We show results worse than the RootSIFT baseline in red, and results better than or equal to RootSIFT in green.} 

\label{fig:mainresults}
\end{figure*}

\begin{figure*}
\begin{center}
\includegraphics[width=\linewidth]{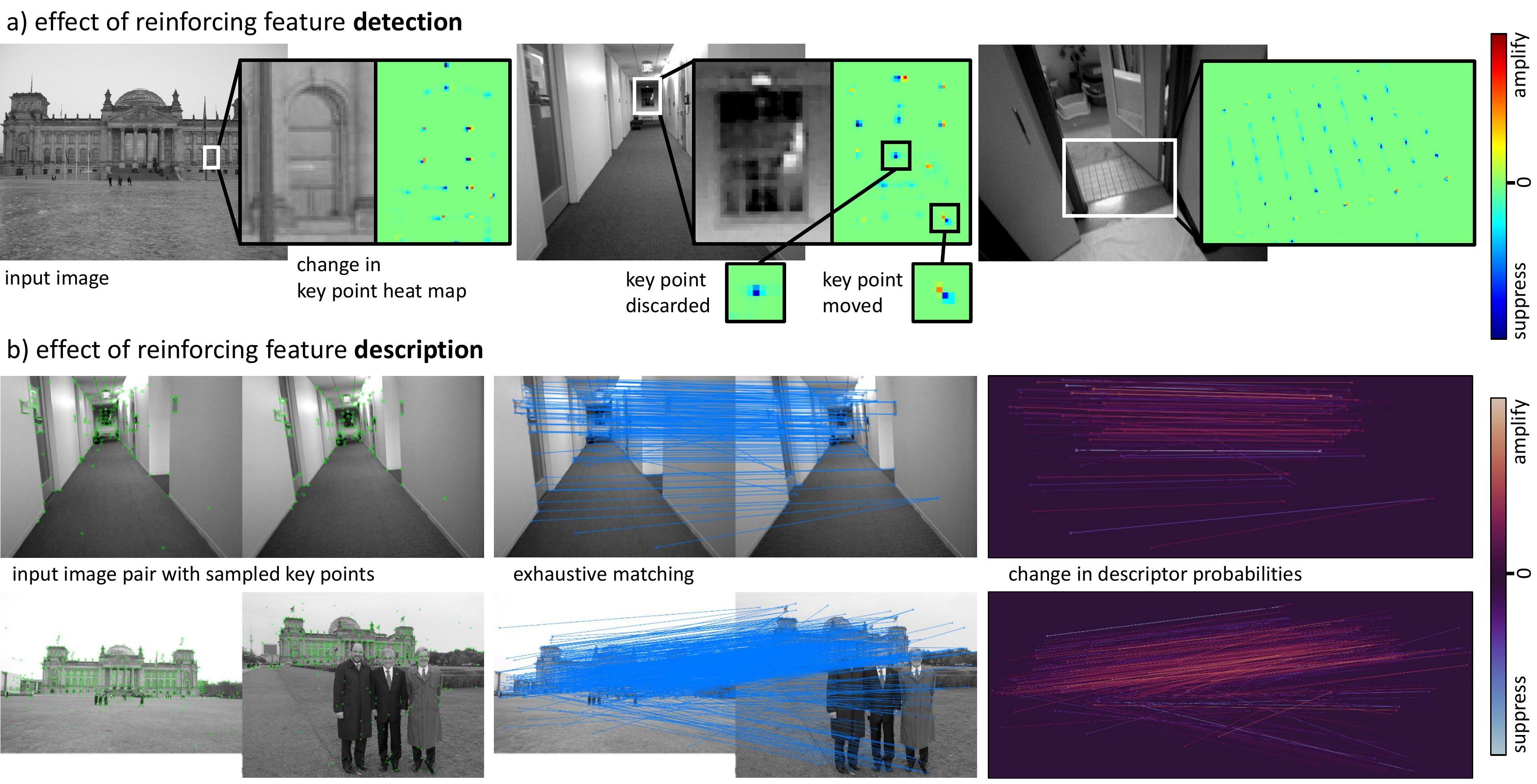}
\end{center}
\vspace{-0.4cm}
   \caption{\textbf{Effect of Training. a)} We visualize the \emph{difference} in key point heat maps predicted by SuperPoint before and after our end-to-end training. Key points which appear blue were discarded, key points with a gradient from blue to red were moved. \textbf{b)} We create a fixed set of matches using (initial) SuperPoint, and visualize the difference in matching probabilities for these matches before and after our end-to-end training. The probability of red matches increased by reducing their descriptor distance relative to all other matches.}
\vspace{-0.4cm}
\label{fig:heatmaps}
\end{figure*}

\paragraph{Discussion.}
We report test accuracy in accordance to Yi \etal. \cite{goodcorr18}, who calculate the pose error as the maximum of rotation and translation angular error.
For each dataset, the area under the cumulative error curve (AUC) is calculated and the mean AUC for outdoor and indoor datasets are reported separately.

Firstly, we train and test our pipeline using a standard RANSAC estimator for essential matrix fitting, see Fig.~\ref{fig:mainresults} a). 
We compare to a state-of-the-art SIFT-based \cite{Lowesift} pipeline, which uses RootSIFT descriptor normalization \cite{rootsift12}.
For RootSIFT, we apply Lowe's ratio criterion \cite{Lowesift} to filter matches where the distance ratio of the nearest and second nearest neighbor is above 0.8.
We also compare to the LIFT feature detector \cite{lift2016}, with and without the learned inlier classification scheme of Yi \etal \cite{goodcorr18} (denoted \emph{InClass}).
Finally, we compare the results of SuperPoint \cite{detone18superpoint} before and after our proposed training (denoted \emph{Reinforced SP}).

Reinforced SuperPoint exceeds the accuracy of SuperPoint across all thresholds, proving that our training scheme indeed optimizes the performance of SuperPoint for relative pose estimation.
The effect is particularly strong for outdoor environments.
For indoors, the training effect is weaker, because large texture-less areas make these scenes difficult for sparse feature detection, in principle.
SuperPoint exceeds the accuracy of LIFT by a large extent, but does not reach the accuracy of RootSIFT.
We found that the excellent accuracy of RootSIFT is largely due to the effectiveness of Lowe's ratio filter for removing unreliable SIFT matches. 
We tried the ratio filter also for SuperPoint, but we found no ratio threshold value that would consistently improve accuracy across all datasets.

To implement a similarly effective outlier filter for SuperPoint, we substitute the RANSAC estimator in our vision pipeline with the recent learning-based NG-RANSAC \cite{Brachmann2019ngransac} estimator.
We train NG-RANSAC for SuperPoint using the public code of Brachmann and Rother \cite{Brachmann2019ngransac}, and with the initial weights for SuperPoint by Detone \etal. \cite{detone18superpoint}.
With NG-RANSAC as a robust estimator, SuperPoint almost reaches the accuracy of RootSIFT, see Fig.~\ref{fig:mainresults}, b).
Finally, we embed both, SuperPoint and NG-RANSAC in our vision pipeline, and train them jointly and end-to-end.
After our training schema, Reinforced SuperPoint matches and slightly exceeds the accuracy of RootSIFT.
Fig.~\ref{fig:mainresults}, c) shows an ablation study where we either update only NG-RANSAC, only SuperPoint or both during end-to-end training. 
While the main improvement comes from updating SuperPoint, updating NG-RANSAC as well allows the robust estimator to adapt to the changing matching statistics of SuperPoint throughout the training process.

\paragraph{Analysis.}
We visualize the effect of our training procedure on the outputs of SuperPoint in Fig.~\ref{fig:heatmaps}.
For the key point heat maps, we observe two major effects.
Firstly, many key points seem to be discarded, especially for repetitive patterns that would result in ambiguous matches.
Secondly, some key points are kept, but their position is adjusted, presumably to achieve a lower relative pose error.
For the descriptor distribution, we see a tendency of reducing the descriptor distance for correct matches, and increasing the descriptor distance for wrong matches.
Quantitative analysis confirms these observations, see Table \ref{tab:stats}.
While the number of key points reduces after end-to-end training, the overall matching quality increases, measured as the ratio of estimated inliers, and ratio of ground truth inliers.

\begin{figure*}
\begin{center}
\includegraphics[width=\linewidth]{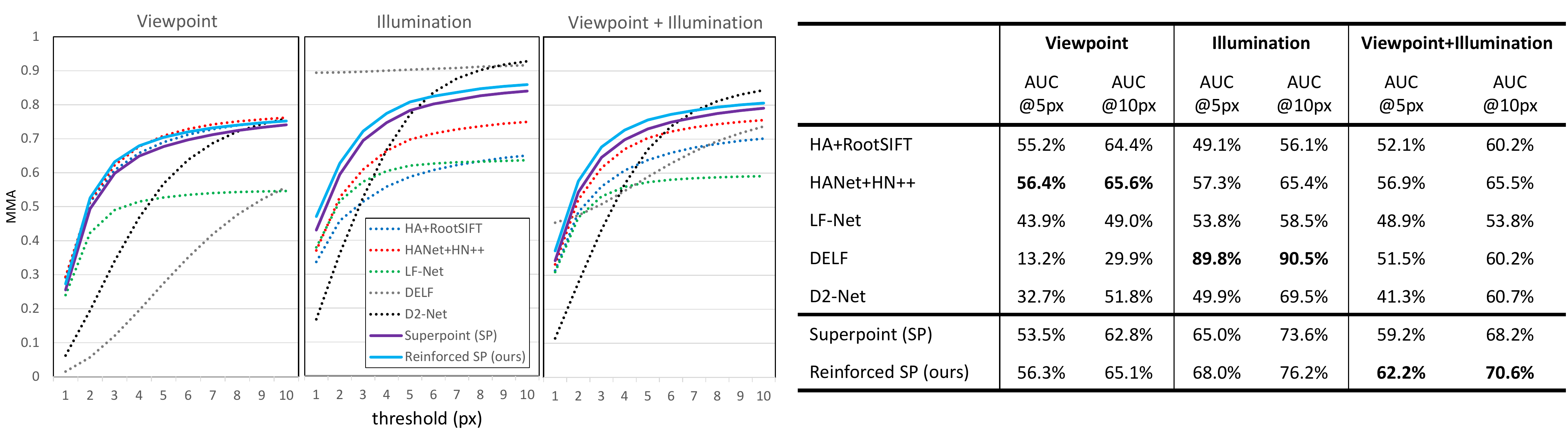}
\end{center}
   \vspace{-0.3cm}
   \caption{\textbf{Evaluation on H-Patches \cite{hpatches2017hpatches}. Left:} We show the mean matching accuracy for SuperPoint before and after being trained for relative pose estimation. Results of competitors as reported in \cite{Dusmanu2019d2net}. \textbf{Right:} Area under the curve (AUC) for the plots on the left.}
   \vspace{-0.3cm}
\label{fig:hpatches}
\end{figure*}

\subsection{Low-Level Matching Accuracy}
\label{sec:exp:hpatches}

We investigate the effect of our training scheme on low-level matching scores.
Therefore, we analyse the performance of Reinforced SuperPoint, trained for relative pose estimation (see previous section), on the H-Patches \cite{hpatches2017hpatches} benchmark. The benchmark consists of 116 test sequences showing images under increasing viewpoint and illumination changes.
We adhere to the evaluation protocol of Dusmanu \etal \cite{Dusmanu2019d2net}.
That is, we find key points and matches between image pairs of a sequence, accepting only matches of mutual nearest neighbours between two images.
We calculate the reprojection error of each match using the ground truth homography. 
We measure the average percentage of correct matches for thresholds ranging from 1px to 10px for the reprojection error.
We compare to a RootSIFT \cite{rootsift12} baseline with a hessian affine detector \cite{Mikolajczyk2004affine} (denoted \emph{HA+RootSIFT}) and several learned detectors, namely HardNet++ \cite{Mishchuk2017hardnet} with learned affine normalization \cite{mishkin2018hanet} (denoted \emph{HANet+HN++}), LF-Net \cite{Ono2018lfnet}, DELF \cite{Noh2017Delf} and D2-Net \cite{Dusmanu2019d2net}.
The original SuperPoint \cite{detone18superpoint} beats all competitors in terms of AUC when combining illumination and viewpoint sequences.
In particular, SuperPoint significantly exceeds the matching accuracy of RootSIFT on H-Patches, although RootSIFT outperforms SuperPoint in the task of relative pose estimation.
This confirms that low-level matching accuracy does not necessarily translate to accuracy in a high-level vision task, see our earlier discussion.
As for Reinforced SuperPoint, we observe an increased matching accuracy compared to SuperPoint, due to having fewer but more reliable and precise key points.

\begin{table}[t]
\begin{tabular}{@{}lcccc@{}}
\toprule
\multicolumn{5}{c}{\textbf{Outdoors}} \\ \midrule
\multicolumn{1}{l|}{} & \multicolumn{1}{l|}{Kps} & \multicolumn{1}{l|}{Matches} & \multicolumn{1}{l|}{Inliers} & \multicolumn{1}{l}{GT Inl.} \\ \midrule
\multicolumn{1}{l|}{SuperPoint (SP)} & \multicolumn{1}{c|}{1993} & \multicolumn{1}{c|}{1008} & \multicolumn{1}{c|}{24.8\%} & 21.9\% \\
\multicolumn{1}{l|}{Reinf. SP (ours)} & \multicolumn{1}{c|}{1892} & \multicolumn{1}{c|}{955} & \multicolumn{1}{c|}{28.4\%} & 25.3\% \\ \midrule
\multicolumn{5}{c}{\textbf{Indoors}} \\ \midrule
\multicolumn{1}{l|}{SuperPoint (SP)} & \multicolumn{1}{c|}{1247} & \multicolumn{1}{c|}{603} & \multicolumn{1}{c|}{13.4\%} & 9.6\% \\
\multicolumn{1}{l|}{Reinf. SP (ours)} & \multicolumn{1}{c|}{520} & \multicolumn{1}{c|}{262} & \multicolumn{1}{c|}{16.4\%} & 11.1\% \\ \bottomrule
\end{tabular}
\vspace{0.1cm}
\caption{Average number of key points and matches found by SuperPoint before and after our training. We also report the estimated ratio of inliers, and the ground truth ratio of inliers.}
\label{tab:stats}
\end{table}

\begin{table}[t]
\begin{tabular}{|l|l|c|c|c|}
\hline
\multicolumn{1}{|c|}{\textbf{Dataset}} & \multicolumn{1}{c|}{\textbf{Method}} & \textbf{\begin{tabular}[c]{@{}c@{}} \# Sparse\\ Points\end{tabular}} & \textbf{\begin{tabular}[c]{@{}c@{}}Track\\ Len.\end{tabular}} & \textbf{\begin{tabular}[c]{@{}c@{}}Repr.\\ Error\end{tabular}} \\ \hline
\multirow{4}{*}{\textbf{\begin{tabular}[c]{@{}l@{}}Fountain\\ (11 img.)\end{tabular}}} & DSP-SIFT & 15k & 4.79 & \textbf{0.41} \\
 & GeoDesc & 17k & \textbf{4.99} & 0.46 \\
 & SuperPoint & \textbf{31k} & 4.75 & 0.97 \\
 & Reinf. SP & 9k & 4.86 & 0.87 \\ \hline
\multirow{4}{*}{\textbf{\begin{tabular}[c]{@{}l@{}}Herzjesu\\ (8 img.)\end{tabular}}} & DSP-SIFT & 8k & 4.22 & \textbf{0.46} \\
 & GeoDesc & 9k & \textbf{4.34} & 0.55 \\
 & SuperPoint & \textbf{21k} & 4.10 & 0.95 \\
 & Reinf. SP & 7k & 4.32 & 0.82 \\ \hline
\multirow{4}{*}{\textbf{\begin{tabular}[c]{@{}l@{}}South\\ Building\\ (128 img.)\end{tabular}}} & DSP-SIFT & 113k & 5.92 & \textbf{0.58} \\
 & GeoDesc & \textbf{170k} & 5.21 & 0.64 \\
 & SuperPoint & 160k & 7.83 & 0.92 \\
 & Reinf. SP & 102k & \textbf{7.86} & 0.88 \\ \hline
\end{tabular}
\vspace{0.1cm}
\caption{Effect of our end-to-end training on a structure-from-motion benchmark. \emph{Reinf.~SP} denotes SuperPoint after being trained for relative pose estimation. Reprojection error is in px.}

\label{tab:colmap}
\end{table}

\vspace{-0.1cm}
\subsection{Structure-from-Motion}
\vspace{-0.1cm}
\label{sec:exp:sfm}

We evaluate the performance of Reinforced SuperPoint, trained for relative pose estimation, in a structure-from-motion (SfM) task.
We follow the protocol of the SfM benchmark of Sch{\"o}nberger \etal \cite{schonberger2017comparative}.
We select three of the smaller scenes from the benchmark, and extract key points and matches using SuperPoint and Reinforced SuperPoint.
We create a sparse SfM reconstruction using COLMAP \cite{schoenberger2016sfm}, and report the number of reconstructed 3D points, the average track length of features (indicating feature stability across views), and the average reprojection error (indicating key point precision). 
We report our results in Table \ref{tab:colmap}, and confirm the findings of our previous experiments.
While the number of key points reduces, the matching quality increases, as measured by track length and reprojection error.
For reference, we also show results for DSP-SIFT \cite{Dong2015DSP} the best of all SIFT variants on the benchmark \cite{schonberger2017comparative},
and GeoDesc \cite{luo2018geodesc}, a learned descriptor which achieves state-of-the-art results on the benchmark.
Note that SuperPoint only provides pixel-accurate key point locations, compared to the sub-pixel accuracy of DSP-SIFT and GeoDesc. 
Hence, the reprojection error of SuperPoint is higher.

\vspace{-0.2cm}
\section{Conclusion}
\vspace{-0.1cm}

We have presented a new methodology for end-to-end training of feature detection and description which includes key point selection, feature matching and robust model estimation.
We applied our approach to the task of relative pose estimation between two images.
We observe that our end-to-end training increases the pose estimation accuracy of a state-of-the-art feature detector by removing unreliable key points, and refining the locations of remaining key points.
We require a good initialization of the network, which might have a limiting effect in training.
In particular, we observe that the network rarely discovers new key points.
Key point locations with very low initial probability will never be selected, and cannot be reinforced.
In future work, we could combine our training schema with importance sampling, for biased sampling of interesting locations.

\vspace{-0.3cm}
\paragraph*{Acknowledgements:}
This project has received funding from the European Social Fund (ESF) and Free State of Saxony under SePIA grant 100299506, DFG Cluster of Excellence CeTI (EXC2050/1 Project ID 390696704), DFG grant 389792660 as part of \href{https://perspicuous-computing.science}{TRR~248}, the European Research Council (ERC) under the European Union’s Horizon 2020 research and innovation programme (grant agreement No 647769), and DFG grant COVMAP: Intelligente Karten mittels gemeinsamer GPS- und Videodatenanalyse (RO 4804/2-1).
The computations were performed on an HPC Cluster at the Center for Information Services and High Performance Computing (ZIH) at TU Dresden.

{\small
\bibliographystyle{ieee}
\bibliography{egbib}
}

\end{document}